\documentclass[10pt]{article} 
\usepackage[accepted]{rlc}

\usepackage{booktabs} 

\usepackage{amsmath,amsfonts,bm}

\def\eqref#1{equation~\ref{#1}}

\def\1{\bm{1}}

\DeclareMathAlphabet{\mathsfit}{\encodingdefault}{\sfdefault}{m}{sl}
\SetMathAlphabet{\mathsfit}{bold}{\encodingdefault}{\sfdefault}{bx}{n}

\newcommand\blfootnote[1]{%
  \begingroup
  \renewcommand\thefootnote{}\footnote{#1}%
  \addtocounter{footnote}{-1}%
  \endgroup
}

\usepackage{hyperref}
\hypersetup{colorlinks,linkcolor={blue},citecolor={magenta},urlcolor={red}}

\usepackage{graphicx}
\usepackage{subfigure}
\usepackage{wrapfig}

\usepackage{algorithm}
\usepackage{algpseudocode}

\usepackage{bbm}

\usepackage{hyperref}
\usepackage{cleveref}
\usepackage{url}

\title{On the consistency of hyper-parameter selection in value-based deep reinforcement learning}

\author{Johan Obando-Ceron\(^{*1,2,3}\), João G.M. Araújo\(^{*3}\), Aaron Courville\(^{1,2}\), \\\textbf{Pablo Samuel Castro\(^{1,2,3}\)
}\\\\
Mila - Québec AI Institute\(^{1}\) \\
Universit\'e de Montr\'eal\(^{2}\) \\
Google DeepMind\(^{3}\)  \\
}

\begin{document}

\maketitle
\blfootnote{*Authors contributed equally. Correspondence to \texttt{jobando0730@gmail.com},\texttt{[joaogui,psc]@google.com}}

\begin{abstract}
Deep reinforcement learning (deep RL) has achieved tremendous success on various domains through a combination of algorithmic design and careful selection of hyper-parameters. Algorithmic improvements are often the result of iterative enhancements built upon prior approaches, while hyper-parameter choices are typically inherited from previous methods or fine-tuned specifically for the proposed technique. Despite their crucial impact on performance, hyper-parameter choices are frequently overshadowed by algorithmic advancements. This paper conducts an extensive empirical study focusing on the reliability of hyper-parameter selection for value-based deep reinforcement learning agents, including the introduction of a new score to quantify the consistency and reliability of various hyper-parameters. Our findings not only help establish which hyper-parameters are most critical to tune, but also help clarify which tunings remain {\em consistent} across different training regimes.

\end{abstract}

\section{Introduction}
\label{sec:introduction}

Sequential decision making is generally considered an essential ingredient for generally capable agents. The ability to plan ahead and adapt to changing circumstances is synonymous with the concept of {\em agency}. For decades, the field of reinforcement learning (RL) has worked on developing methods, or agents, for precisely this purpose. This research has borne impressive results, such as developing agents which can play difficult Atari games \citep{mnih2015humanlevel}, control stratospheric balloons \citep{Bellemare2020AutonomousNO}, control a tokamak fusion reactor \citep{Degrave2022MagneticCO}, among others. These are all examples of {\em deep reinforcement learning} (DRL), which combines the theory of reinforcement learning with the expressiveness and flexibility of deep neural networks.

The success of these methods built on years of academic research, where novel algorithms and techniques were introduced and showcased on academic benchmarks such as the ALE \citep{bellemare2012ale}, MuJoCo \citep{todorov2012mujoco}, and others. These benchmarks typically consist of a suite of environments that have varied transition and reward dynamics. Their common usage provides us with a familiarity which affords us a sense of interpretability, a consistency in evaluation that grants us a sense of reliability, and their variety yields a sense of generalizability. Unfortunately, this promise often fails to materialize: their reliability has been brought into question by numerous works which demonstrate their fickleness \citep{Henderson2017DeepRL,agarwal2021deep}, while there is a general sentiment that researchers have ``overfit’’ to these benchmarks, bringing into question their generalizability. A critical aspect to these challenges is the difficulty in training neural networks in an RL setting \citep{ostrovski2021the,lyle2022learning,sokar2023dormant}.

Although the successes above built on prior methods, they were not taken ``as is’’: it took large teams of researchers many months and lots of compute to adapt prior work to their specific problem. These adaptations include changes to the network architectures, designing reward functions to induce the desired behaviours, and careful tuning of the many hyper-parameters. This last point is indeed {\em essential} to the success of any DRL method: improper hyper-parameter choices can cause a theoretically sound method to drastically underperform, while careful hyper-parameter selection can dramatically increase the performance of an otherwise sub-optimal method.

As an example of this dichotomy, we examine how DER \citep{hasselt19when}, a method that has become a common baseline for the Atari $100$k benchmark \citep{kaiser2020modelbased}, came to be. DQN, considered to be the start of the field of DRL research, was introduced by showcasing its super-human performance on the ALE \citep{bellemare2012ale}, a suite of $57$ Atari $2600$ games. This suite became one of the most popular benchmarks on which to evaluate new methods over $200$ million environment frames\footnote{See \citep{machado2018revisiting} for more details on ALE evaluation standards.}. A few years later, when \citet{kaiser2020modelbased} introduced the SiMPLe algorithm as a sample-efficient method, they argued for evaluating it only on $100$k agent actions\footnote{The standard for ALE agents is to use frame-skipping, where $4$ environment frames occur for every agent action. This results in frustratingly confusing nomenclature, as $200$M is specified in environment frames (or $500$k agent actions), while $100$k is specified in agent actions (or $400$k environment frames).} with a subset of $26$ games, so as to properly test the sample-efficiency of new methods. The authors demonstrated that their proposed method outperformed Rainbow \citep{Hessel2018RainbowCI}, the state-of-the-art method of the time. In response, \citet{hasselt19when} introduced Data Efficient Rainbow (DER), which outperformed SiMPLe even though it was the same Rainbow algorithm, but {\em with a careful tuning of the hyper-parameters for the $100$k training regime}.

One could argue that the hyper-parameters of Rainbow were overly-tuned to the $200$M benchmark, while the hyper-parameters of DER were overly-tuned to the $100$k benchmark. More importantly, what this story highlights is that, despite careful evaluation it is quite likely that a new method {\em will not work as intended when deployed on a different environment from which it was trained on}, and that a significant  amount of hyper-parameter tuning will be necessary. This flies in the face of the supposed generalizability of DRL academic research, and makes it difficult for groups without large computational budgets to successfully apply prior work to applied problems.

It thus behooves the community to develop a better understanding of the {\em transferability} and {\em consistency} of hyper-parameter selection across different training regimes, and to build a better shared understanding of the relative importance of the many possible hyper-parameters to tune. In this work, we take a stride towards this by conducting an exhaustive empirical investigation of the various hyper-parameters affecting DRL agents. We focus our attention on two value-based agents developed for the Atari $100$k suite: DER mentioned above, and DrQ($\epsilon$), a variant of DQN that was optimized for the $100$k suite. Although developed for the $100$k suite, we also train these agents for $40$M million environment frames. Our intent is to examine the transferability of various hyper-parameter choices across different training regimes. Specifically, we investigate:
{\bf Across data regimes:} Do hyper-parameters selected in the $100$k regime work well in a larger data regime? {\bf Across agents:} Do hyper-parameters selected for one agent work well in another? {\bf Across environments:} Do hyper-parameters tuned in one set of environments work well in others?

In total, we investigated $12$ hyper-parameters with different values for $2$ agents over $26$ environments, each for $5$ seeds, resulting in a total of $108$k independent training runs. This breadth of experimentation results in an overwhelming amount of data which complicates their analyses. We address this challenge in two ways: \textit{(i)} We introduce a new score which provides us with an aggregate value for the considerations mentioned above. \textit{(ii)} We provide an interactive website where others may easily navigate the large number of experimental figures we have generated.

The score provides us with a high-level overview of our findings, while the website grants us a fine-grained mechanism to analyze the results. We hope this effort provides the community with useful tools so as to develop not just better DRL algorithms, but better methodologies to evaluate their interpretability, reliability, and generalizability.
\section{Background}
\label{sec:brackground}


The field of reinforcement learning studies algorithms for sequential decision-making problems. In these settings, an algorithm (or agent) interacts with an {\em environment} by transitioning between {\em states} and making action choices at discrete timesteps; the environment responds to each action by (possibly) changing the agent's state and yielding a numerical reward or cost. The goal of the agent is to maximize the cumulative rewards (or minimize the cost) throughout its lifetime.
This is typically formalized as a Markov decision process (MDP) \citep{puterman2014markov} $\langle \mathcal{X}, \mathcal{A}, \mathcal{P}, \mathcal{R}, \gamma \rangle$, where $\mathcal{X}$ is the set of states, $\mathcal{A}$ is the set of available actions, $\mathcal{P}:\mathcal{X}\times\mathcal{A}\rightarrow \Delta(\mathcal{X})$\footnote{$\Delta(X)$ denotes a distribution over the set $X$.} is the transition function, $\mathcal{R}:\mathcal{X}\times\mathcal{A}\rightarrow\mathbb{R}$ is the reward function, and $\gamma\in [0, 1)$ is a discount factor. An agent's behaviour is formalized by a policy $\pi:\mathcal{X}\rightarrow\Delta(\mathcal{A})$, whose {\em value} from any state $x\in\mathcal{X}$ is given by the Bellman recurrence 
$V^{\pi}(x) := \mathbb{E}_{a\sim\pi(x)}\left[\mathcal{R}(x, a) + \gamma \mathbb{E}_{x'\sim\mathcal{P}(x, a)}V^{\pi}(x')\right]$. $Q$-functions allow us to measure the value of taking any action $a\in\mathcal{A}$ from a state $x\in\mathcal{X}$ and following $\pi$ afterwards: $Q^{\pi}(x, a) := \mathcal{R}(x, a) + \gamma \mathbb{E}_{x'\sim\mathcal{P}(x, a)} V^{\pi}(x')$. 
A policy $\pi^*$ is considered optimal if for any policy $\pi$, $V^* := V^{\pi^*} \geq V^{\pi}$.

Solving for the equations discussed above would require access to both $\mathcal{R}$ and $\mathcal{P}$, which are usually unknown. Instead, RL typically assumes the agent has access to transitions $\tau := (x, a, r, x')\in\mathcal{X}\times\mathcal{A}\times\mathbb{R}\times\mathcal{X}$, arising from interactions with the environment. Given such a transition,  $Q$-learning \citep{Watkins1992qlearning} updates its estimate of $Q$ via: $Q_{t+1}(x, a) \leftarrow  Q_t(x, a) + {\alpha} TD(Q, \tau)$, where {$\alpha$} is a learning rate and $TD$ is the {\em temporal-difference error}, given by $TD(Q_t, \tau) := r + \gamma \max_{a' \in\mathcal{A}} Q_t(x', a') - Q_t(x, a)$. If the state and action spaces are small, one can store all the $Q$-values in a table of size $|\mathcal{X}|\times |\mathcal{A}|$. For most problems of interest, however, state spaces are very large (and possibly infinite). In these cases, one can use a function approximator, such as a neural network, parameterized by $\theta$: $Q_\theta\approx Q$. Indeed, in order to achieve super-human performance on the Arcade Learning Environment (ALE) \citep{bellemare2012ale}, \citet{mnih2015humanlevel} used a neural network consisting of three convolutional layers (Conv layers), followed by two multi-layer perceptrons (Dense layers) with $|\mathcal{A}|$ outputs in the final layer (representing the $Q$-value estimates for each action). With the exception of the final layer, a ReLU non-linearity follows each layer.

Updating $Q_{\theta}$ thus corresponds to updating the parameters $\theta$, which may be done by using optimization algorithms such as Adam \citep{kingma15adam} to minimize the temporal-difference error. At a high-level, this yields an update of the form: $\theta_{t+1} \leftarrow \theta_t + \alpha\nabla_{\theta_t} \mathbb{E}_{\tau\sim\mathcal{D}} TD(Q_{\theta_t}, \tau)$.
The expectation can be approximated using a batch of $m$ transitions drawn from a distribution $\mathcal{D}$, which can be computed efficiently on specialized hardware such as GPUs and TPUs. Additionally, \citet{mnih2015humanlevel} argued that using $\bar{\theta}$, a less-frequently updated copy of the parameters, when computing TD helps with training stability. A common approach introduced by \citet{mnih2015humanlevel} is to clip the rewards at $(-1, 1)$. The TD term thus becomes:
  $TD(Q_{\theta}, \tau) := clip(r, (-1, 1)) + \gamma\max_{a'\in\mathcal{A}}Q_{\bar{\theta}}(x', a') - Q_{\theta}(x, a)$.

Although DQN benchmarked on the 57 ALE games with the same set of hyper-parameters, \citet{anschel2017averageddqn} 
demonstrated that in some environments it can result in degraded performance.
A number of papers have proposed improvements to increase stability and performance, which \citet{Hessel2018RainbowCI} combined into a single agent they called \emph{Rainbow}. Specifically, they combined DQN with double Q-learning \citep{hasselt2015doubledqn}, prioritized experience replay \citep{Schaul2016PrioritizedER}, dueling networks \citep{wang16dueling}, multi-step learning \citep{sutton88learning}, noisy nets \citep{fortunato18noisy}, and distributional reinforcement learning \citep{Bellemare2017ADP}.

\section{THC Score}
\label{sec:thc_metric}

Statistical metrics play a crucial role in assessing and evaluating the performance of DRL algorithms. 
They provide valuable insights into the strengths and weaknesses of different approaches, guiding researchers and practitioners in the development of more effective reinforcement learning systems.
For example, some the metrics focus on the mean reward obtained by an agent per time step (Average Reward), the percentage of episodes in which the agent achieves a predefined goal or task (success rate) among others \citep{agarwal2021deep, chan2020measuring, Henderson2017DeepRL}. 

Measuring the transferability/consistency of hyper-parameters in DRL is challenging, as existing metrics fall short in capturing the nuanced aspects of how well hyper-parameter settings generalize across different environments or agents. Developing such a metric would enhance the ability to systematically compare and select hyper-parameter configurations that exhibit robust performance across a range of application domains.

To understand the consistency of hyper-parameters we focus on their ranking consistency across experimental settings. Put another way: if a given hyper-parameter value is optimal/pessimal in a setting, is it still optimal/pessimal in another? And so we analyse, for each hyper-parameter, whether their values lead to the same ranking order for different experimental settings, where the ranking is on final performance. 

We compute ranking agreement for three setups: 
{\bf $1$) Varying algorithms} while keeping the environment and data regime fixed (e.g. when proposing a new value-based algorithm but not having enough compute to run a comprehensive hyper-parameter search). {\bf $2$) Varying environments} while keeping the algorithm and data regime fixed (e.g. when using a state of the art algorithm in a new domain).
{\bf $3$) Varying data regimes} while keeping the environment and algorithm fixed (e.g. when adapting a new algorithm to a new data regime \citep{hasselt19when}).
Concretely, our desire is to have a metric that yields a high value score would indicate that the hyper-parameter in question is {\em important}, in the sense that it will likely require retuning; conversely, a low score suggests the hyper-parameter value can likely be kept as is.

Kendall's Tau \citep{kendall38measure} and Kendall's W \citep{10.1214/aoms/1177732186} are natural choices, but these metrics were developed for situations where the rankings were based on a single score, instead of a range of possible scores, and they can result in degenerate values when two settings have similar performance or when two settings alternate between optimal and pessimal rankings. For these reasons, we introduce the \textbf{T}uning \textbf{H}yperparameter \textbf{C}onsistency ({\bf THC}) score. Consider a set of $n$ hyper-parameters $\lbrace H_1,\ldots,H_n\rbrace$, each with its set of values $\lbrace\lbrace h_{11},h_{12},\ldots,h_{1m_1}\rbrace, \ldots,\lbrace h_{n1},h_{n2},\ldots,h_{nm_n}\rbrace\rbrace$ (e.g. hyper-parameter $H_i$ has $m_i$ values). The THC score involves three computations: (i) rankings for each hyper-parameter setting (\autoref{alg:computeRankings}); (ii) normalized peak-to-peak value for each hyper-parameter setting (Eqn.~\ref{eqn:ptp} below); and (iii) overall THC score for the hyper-parameter (see Eqn.~\ref{eqn:thc} below).

If we run multiple independent runs for each hyper-parameter setting $h_{ij}$, we can compute the mean $\mu_{ij}$ and standard deviation $\sigma_{ij}$ for these runs\footnote{One may also use confidence intervals instead of standard deviations.}. For each hyper-parameter setting $h_{ij}$ we then compute an initial ranking $r'_{ij}$ based on the upper bound ($\mu_{ij}+\sigma_{ij}$), with the lower bound ($\mu_{ij}-\sigma_{ij}$) used to break ties. We then define a set containing hyper-parameter settings with overlapping values:
\begin{align*}
    I_{ij} := \{k \vert (\mu_{ij} - \sigma_{ij} \leq \mu_{ik} + \sigma_{ik} &\text{ and } \mu_{ij} - \sigma_{ij} \geq \mu_{ik} - \sigma_{ik}) \\ &\text{ or } \\ \break (\mu_{ij} + \sigma_{ij} > \mu_{ik} - \sigma_{ik} &\text{ and } \mu_{ij} + \sigma_{ij} < \mu_{ik} + \sigma_{ik}) \}
\end{align*}

\begin{algorithm}[!t]
\caption{Compute rankings}\label{alg:computeRankings}
\begin{algorithmic}[1]
\Require Multiple runs for various settings of hyper-parameter $H_i$: $\lbrace h_{i1},h_{i2},\ldots,h_{im_i}\rbrace$, aggregate metrics $\mu_i$: $\lbrace \mu_{i1},\mu_{i2},\ldots,\mu_{im_i}\rbrace$ and measure of spread $\sigma_i$: $\lbrace \sigma_{i1},\sigma_{i2},\ldots,\sigma_{im_i}\rbrace$
\For{$i$ in $1 \ldots n$}
    \State $r'_{i} = \textrm{argsort}(\mu_i + \sigma_i)$ \Comment{Gets the index of each value as if the array was sorted}
    \State $\mu'_i, \sigma'_i = \mu_i[r'_{i}], \sigma_i[r'_{i}]$ \Comment{Sorted versions of aggregate and spread metrics}
    \For{$j$ in $1 \ldots m_{i}$} 
        \State $u_{j} = \textrm{binary\_search}(\mu'_i - \sigma'_i, \mu_{ij} + \sigma_{ij})$ \Comment{highest rank whose lower bound overlaps with j}
        \State $l_{j} = \textrm{binary\_search}(\mu'_i + \sigma'_i, \mu_{ij} - \sigma_{ij})$ \Comment{lowest rank whose upper bound overlaps with j}
    \EndFor
    \State $\bf{r_{i}} = \frac{u + l}{2}$ \Comment{The average rank in $l_j,l_j+1, \ldots, u_j$ is the average of $l_j$ and $u_j$}
\EndFor

\end{algorithmic}
\end{algorithm}

The final ranking of each hyper-parameter is $r_{ij} = \frac{\sum_{k \in I_{ij}} r
'_{ik}}{\vert I_{ij} \vert}$, 
as \autoref{alg:computeRankings} details. These rankings are for {\em one} training regime; however, as mentioned in the introduction, we are interested in quantifying the {\em consistency} of a hyper-parameter $H$ across varying training regimes. Consider four training regimes $A, B, C, D$, and let $\lbrace \mathfrak{R}^A,\ldots,\mathfrak{R}^D\rbrace$ denote their respective rankings. For each hyper-parameter value $h_x\in H$ we compute its normalized ``peak-to-peak''\footnote{Inspired by numpy's peak-to-peak function numpy.ptp \citep{harris2020array}.} value $\overline{\textrm{ptp}}$, which quantifies its variance in ranking, as follows: First compute the $\textrm{ptp}$ value $\textrm{ptp}(h_x) = \max\left(\lbrace \mathfrak{R}^A(h_x),\ldots,\mathfrak{R}^D(h_x)\rbrace\right) - \min\left(\lbrace \mathfrak{R}^A(h_x),\ldots,\mathfrak{R}^D(h_x)\rbrace\right)$, then normalize so they're in [0, 1]:
\begin{align}
    \overline{\textrm{ptp}}(h_x) = \frac{\textrm{ptp}(h_x)}{|H| - 1}
    \label{eqn:ptp}
\end{align}

Notably, hyper-parameter settings that have consistent rankings across training regimes will have a normalized $\textrm{ptp}$ value of zero. Finally, the $\textrm{THC}$ score for hyper-parameter $H$ is defined as:
\begin{align}
    \textrm{THC}(H) = \frac{\sum_{h_x\in H}\overline{\textrm{ptp}}(h_x)}{|H|}.
    \label{eqn:thc}
\end{align}

This score will result in low values for hyper-parameters whose varying settings have consistent ranking across various training regimes, and high values when these rankings vary. Intuitively, {\em hyper-parameters with high values will most likely require re-tuning when switching training regimes}. See \autoref{sec:appendixTHC} for more examples of computing the score, as well as the source code provided with this submission.

\section{Hyper-parameters considered} 
\label{sec:hyper-parameter_selection}

We describe the set of hyper-parameters explored in this work, with the values used for each listed in \autoref{sec:list_hyperparameters}. Unless otherwise specified, these are examined for both Conv and Dense layers.

{\bf Activation functions:} 
Non-linear activation functions are a fundamental part of neural networks, as their removal effectively turns the network into a linear function approximator.
While various activation functions have been proposed \citep{devlin2019bert, Elfwing2018SigmoidWeightedLU, 10.5555/3305381.3305478}, there have been few works comparing their performance \citep{Shamir2020SmoothAA}; to the best of our knowledge, there are no previous examples of such a comparison in the RL setting.

{\bf Normalization: }
Normalization plays an important role in supervised learning \citep{tan2020efficientnet, xie2017aggregated} but is relatively rare in deep reinforcement learning, with a few exceptions \citep{gogianu2021spectral, bhatt2019crossnorm, arpit2019initialize, alphaZero}. We explore {\em batch normalization} \citep{ioffe2015batch} and {\em layer normalization} \citep{ba2016layer}.

{\bf Network capacity: } 
``Scaling laws'' have been central to the growth of capabilities in large language/vision models, but have mostly eluded reinforcement learning agents, with a few exceptions \citep{schwarzer23a, taiga2022investigating, farebrother2022proto,obando2024mixtures,obandoceron2024pruned,farebrother2024stop}. 
To investigate the impact of network size, we vary the {\em depth} (e.g. the number of hidden layers) and the {\em width} (e.g. the number of neurons of each hidden layer).

{\bf Optimizer hyper-parameters: }
\label{sec:optimizerHypers}
We explore three hyper-parameters of Adam \citep{kingma15adam}, which has become the standard optimizer used by most: {\em learning rate}, {\em epsilon} and {\em weight decay}.
\emph{Learning rate} determines the step size at which the algorithm adjusts the model's parameters during each iteration.
$\epsilon$ represents a small constant value that is added to the denominator of the update rule to avoid numerical instabilities.
\emph{Weight decay} adds a penalty term to the loss function during training that discourages the model from assigning excessively increasing weight magnitudes.

{\bf $\epsilon$-greedy exploration: } 
$\epsilon$-greedy exploration is a simple and popular exploration technique which picks actions greedily with probability $1-\epsilon$, and a random action with probability $\epsilon$. Traditionally, experiments on the ALE use a linear decay strategy to decay $\epsilon$ from $1.0$ to its target value.

{\bf Reward clipping: } 
Most ALE experiments clip rewards at $(-1, 1)$ \citep{mnih2015humanlevel}.

{\bf Discount factor: } 
The multiplicative factor $\gamma$ discounts future rewards and its importance has been observed in a number of recent works \citep{amit2020discount, hessel19inductive, gelada2019off, vanseijen2019using, francoislavet2016discount,schwarzer23a}.

{\bf Replay buffer: }  
DRL agents  store past experiences in a replay buffer, to sample from during learning. The {\em replay capacity} parameter refers to the amount of data experiences stored in the buffer. 
It is common practice to only begin sampling from the replay buffer when a minimum number of transitions have been stored, referred to as the {\em minimum replay history}.

{\bf Batch size: } 
The number of stored transitions that are sampled for learning at each training step.

{\bf Update horizon: }
Multi-step learning \citep{sutton88learning} computes the temporal difference error using multi-step transitions, instead of a single step. DQN uses a single-step update by default, whereas Rainbow chose a 3-step update \citep{Hessel2018RainbowCI}. The update horizon has been argued to trade-off between the bias and the variance of the return estimate \citep{biasandvariance_kea}.

{\bf Target Update periods: }
Value based agents often employ an online and a {\em target} Q-network, the latter which is updated less frequently by directly syncing (or Polyak-averaging) from the online network; the {\em target updated period} determines how frequently this occurs.

{\bf Update periods: }
The online network parameters are updated after every {\em update period} environment steps, with a value of $4$ used in standard ALE training.

{\bf Number of atoms: } 
In distributional reinforcement learning \citep{Bellemare2017ADP}, the output layer predicts the distribution of the returns for each action $a$ in a state $s$, instead of the mean $Q^{\pi}(s, a)$. A popular approach is to model the return as a categorical distribution parameterized by a certain number of 'atoms' over a pre-specified support.

\begin{figure}[!t]
    \centering
  \includegraphics[width=\linewidth]{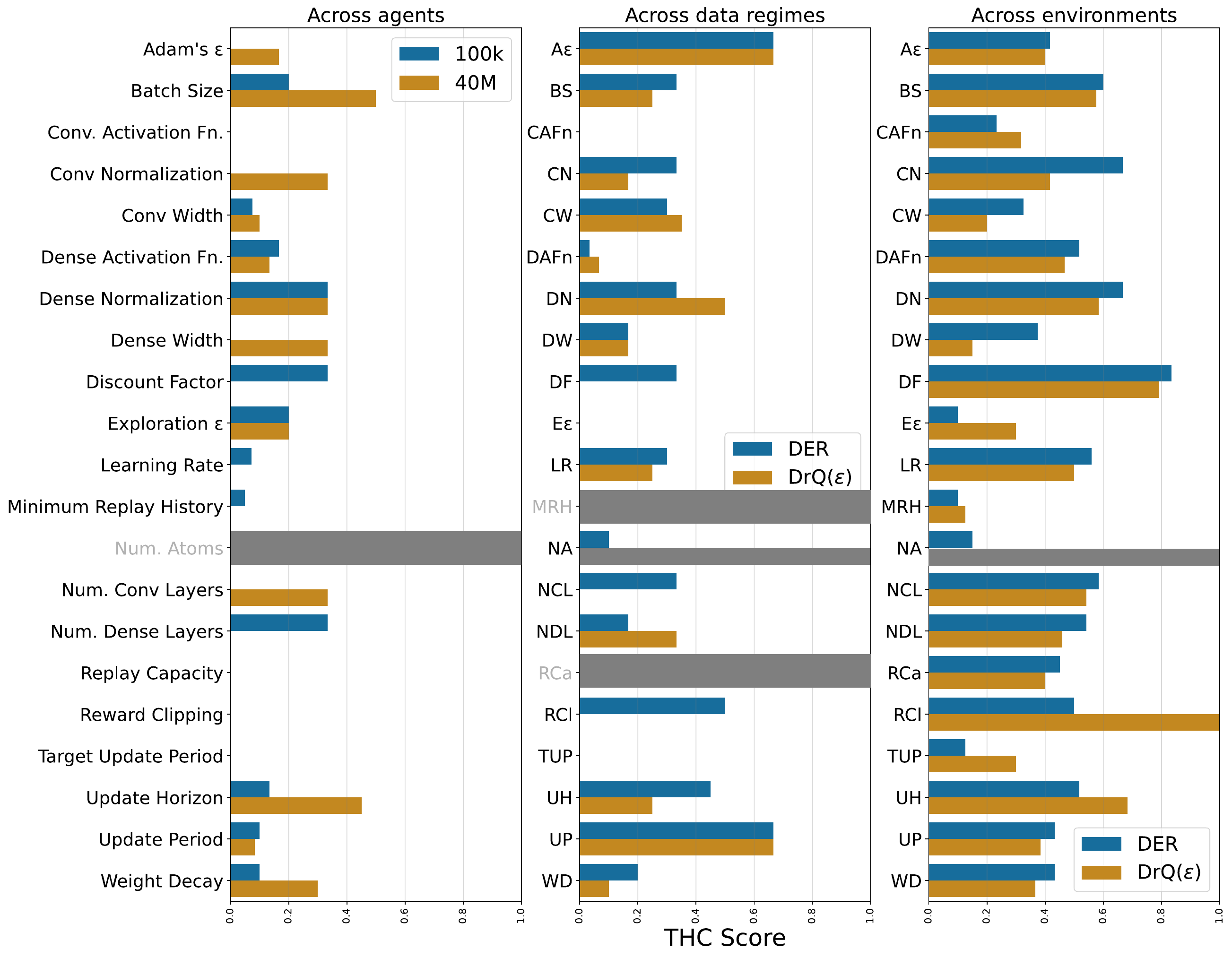}%
    \caption{Tuning hyper-parameter Consistency (THC Score, see \cref{sec:thc_metric}) evaluated across agents (\textbf{left panel}), data regimes (\textbf{center panel}), and environments  (\textbf{right panel}). Different colors indicate different data regimes (left panel) and different agents (center and right panels); grey bars/titles indicate hyper-parameters which are not comparable across the considered transfer settings.
    \label{fig:this_score_all}%
    }%
\end{figure}

\section{Experimental results} 
\label{exp_results}
As mentioned in the introduction, there already exist two data regimes for evaluating agents on the ALE suite: the (low-data regime) $100$k \citep{kaiser2020modelbased} and the original $200$M benchmark \citep{mnih2015humanlevel}. The $100$k benchmark includes only $26$ games from the original suite, so we focus on these for our evaluation. For computational considerations, we follow \citet{graesser2022state} and use $40$M million environment frames as our large-data regime.
We use the settings of DrQ($\epsilon$) (introduced by \citet{agarwal2021deep} as an improvement over the DrQ of \citet{yarats2021image}), and 
Data Efficient Rainbow (DER) introduced by \citet{hasselt19when}. All experiments were run on a Tesla P100 GPU and took around $2$-$4$ hours ($100$k) and $1$-$2$ days ($40$M) per run.
Both algorithms are implemented in the Dopamine library \citep{castro18dopamine}. Since the $100$k setting is cheaper, we evaluated a larger set of hyper-parameter values there and manually picked the most informative subset for running in the $40$M setting. For all our experiments we ran 5 independent seeds and followed the guidelines suggested by \citet{agarwal2021deep} for more statistically meaningful comparisons. Specifically, we computed aggregate human-normalized scores and report interquantile mean (IQM) with $95\%$ stratified bootstrap CIs. 

In \autoref{fig:this_score_all} we present the computed THC score for all the hyper-parameters discussed in \cref{sec:hyper-parameter_selection}, and we discuss their consistency across agents in Section~\ref{sec:acrossAlgorithms}, across data regimes in Section~\ref{sec:acrossData}, and  across environments in Section~\ref{sec:acrossEnvironments}. More detailed discussions are provided in \autoref{sec:finerGrainedExperiments} and a set of interesting findings in \autoref{sec:imf}. It is worth recalling that higher THC scores indicate less consistency, which suggests a likely need to re-tune the respective hyper-parameters when changing training configurations.

\subsection{Optimal hyper-parameters mostly Transfer Across Agents}
\label{sec:acrossAlgorithms}
\begin{figure}[!h]
    \centering
  \includegraphics[width=0.8\linewidth]{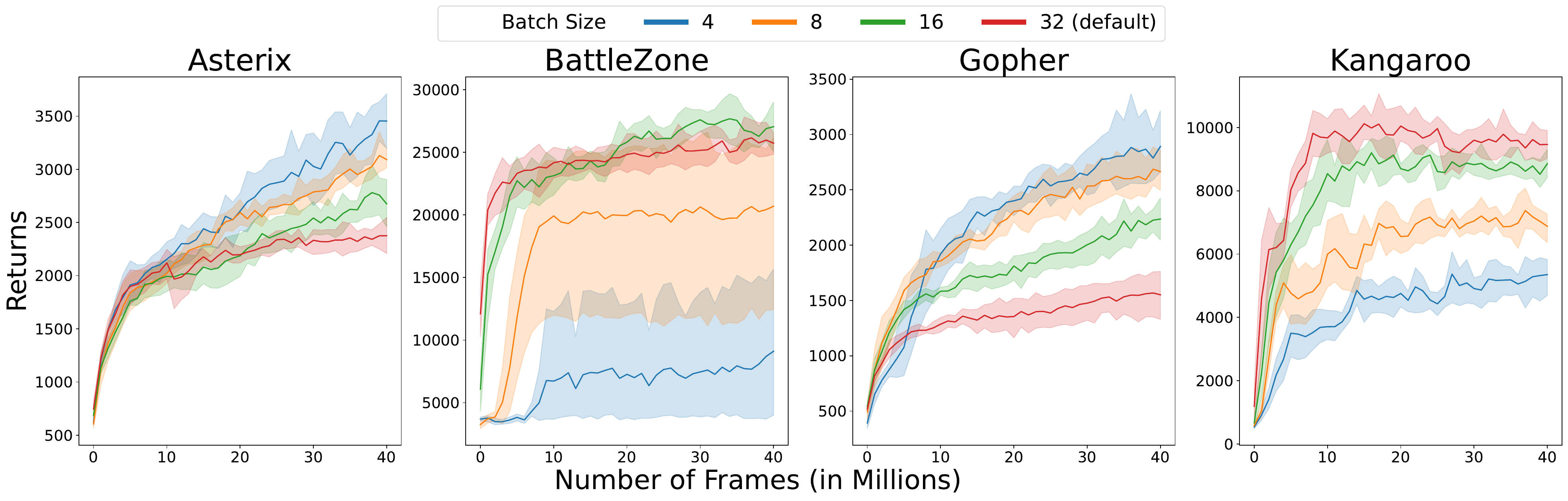}
  \includegraphics[width=0.8\linewidth]{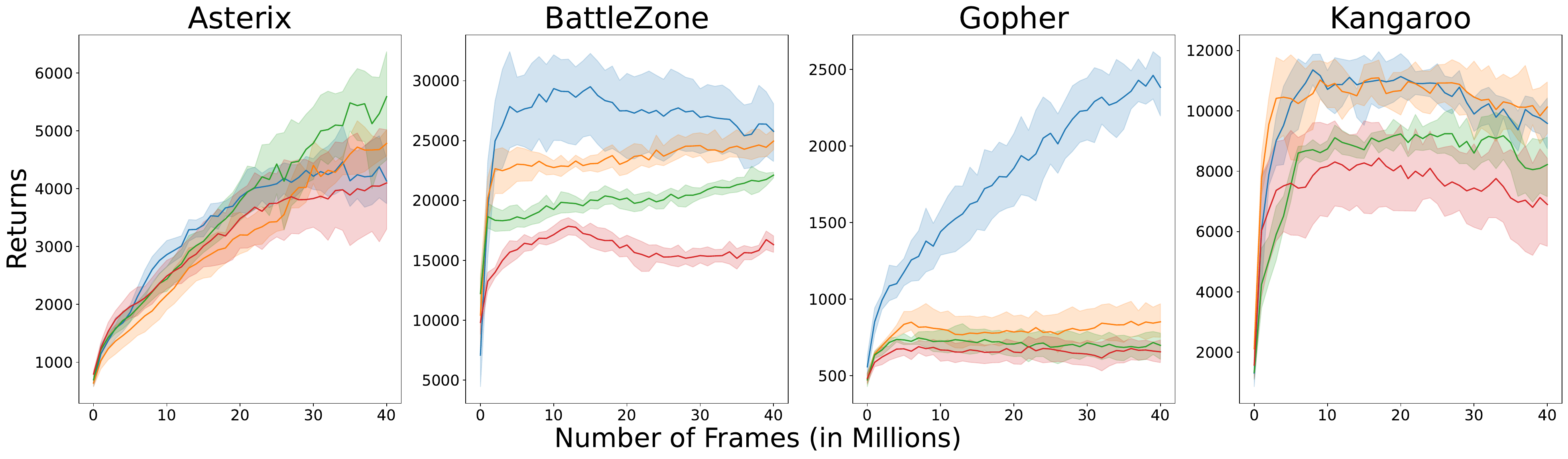}
    \caption{\textbf{Measured returns with varying batch size} for DrQ($\epsilon$) (top) and DER (bottom) at $40$M environment frames for four representative games, demonstrating that the ranking of the hyper-parameter values can drastically change from one game to the next. All results averaged over $5$ seeds, shaded areas represent $95\%$ confidence intervals.
    }%
    \label{fig:drq_eps_batch_sizes}%
\end{figure}

We find that optimal hyper-parameters for DrQ($\epsilon$) agree quite often with DER, which is somewhat expected given that they're based on the same classical RL algorithm of Q-learning, and have the same number of updates in the same environments. Looking at THC values between the two agents for different data regimes we see that all values are below $0.5$, and in the $100$k regime tend to be even lower. Nevertheless, comparing the results of the two rows in \cref{fig:drq_eps_batch_sizes,fig:per_game} demonstrate that there can still be strong differences between the two. In the $40$M regime, the hyper-parameters with the highest THC are batch size and update horizon, consistent with the findings of \cite{obandoceron2023small}, where these two hyper-parameters proved crucial to boosting agent performance.

\begin{figure}[!t]
    \centering
  \includegraphics[width=0.8\linewidth]{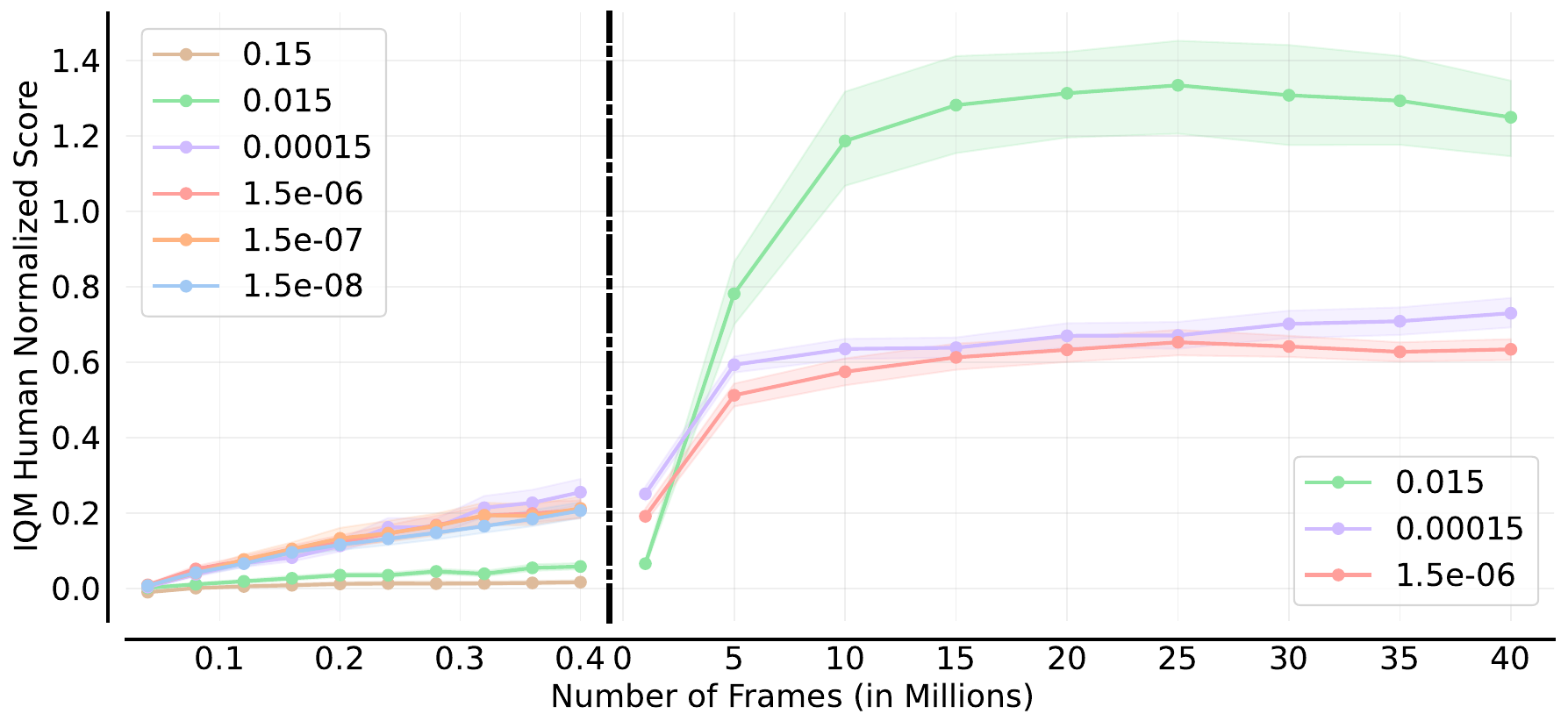}%
    \caption{
     \textbf{Measured IQM of human-normalized scores on the $26$ $100$k benchmark games, with varying Adam's $\epsilon$} for DER. We evaluate performance at 100k agent steps (or 400k environment frames), and at $40$ million environment frames. The ordering of the best hyper-parameters switches between the two data regimes.
    }
    \label{fig:der_adam_eps}
\end{figure}

\subsection{Optimal hyper-parameters mostly do not Transfer Across Data Regimes}
\label{sec:acrossData}
We find that optimal hyper-parameters for Atari 100k mostly do not transfer once you move to 40M updates, showing that even when keeping algorithms and environment constant one may still need to tune hyper-parameters should they change the amount of data their agent can train on. Of the hyper-parameters considered, {\em Adam's $\epsilon$} and {\em update period} seem to be the most critical to re-tune (see \autoref{fig:der_adam_eps} for results on DER for Adam's $\epsilon$). The results with Adam's $\epsilon$ are surprising, as the purpose of this hyper-parameter is mostly for numerical stability. The update period (as well as the update horizon) results are consistent with what is done in practice between these two data regimes (e.g. Rainbow uses an update period of $4$ and an update horizon of $3$, while DER uses $1$ and $10$, respectively).

\subsection{Optimal hyper-parameters do not Transfer Across Environments}
\label{sec:acrossEnvironments}
Our experiments show that hyper-parameters that perform well on some games lead to lackluster final performance in others. Indeed, in \autoref{fig:this_score_all} we can see that the THC score is highest when evaluating across environments. This strongly suggests that, when using an existing agent in a new environment, most of the hyper-parameters would need extra tuning.
\autoref{fig:drq_eps_batch_sizes} displays the results when varying batch size, where we can see that the rankings can sometimes be complete opposites across games (compare Kangaroo and Gopher).

\section{A web-based appendix} 
\label{web_results}
We have run an extensive number of experiments (around 108k) for this work, which would render a traditional appendix unwieldy. Instead, we provide an interactive website\footnote{Website available at \href{https://consistent-hparams.streamlit.app/}{\emph{https://consistent-hparams.streamlit.app/}}.} which facilitates navigating the full set of results\footnote{Website repository at \href{https://github.com/joaogui1/Consistent-Website?tab=readme-ov-file}{\emph{https://github.com/Consistent-Website}}.}. Presenting empirical research results in this manner offers a range of benefits that enhance accessibility, engagement, and comprehension. 
This dynamic presentation allows readers to more easily make comparisons over different games, agents, and parameters.

\begin{figure}[!t]
    \centering
   \includegraphics[width=\textwidth]{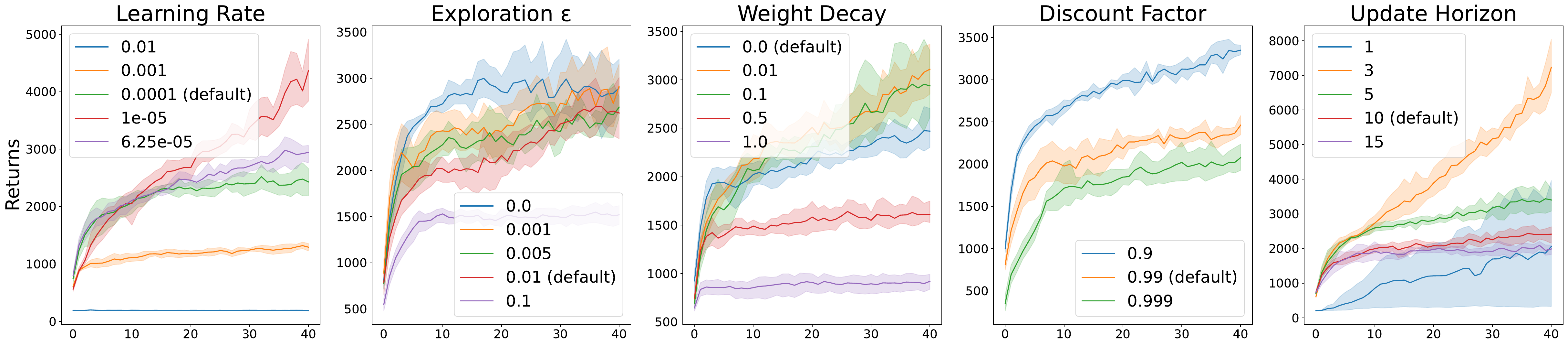}
   \includegraphics[width=\textwidth]{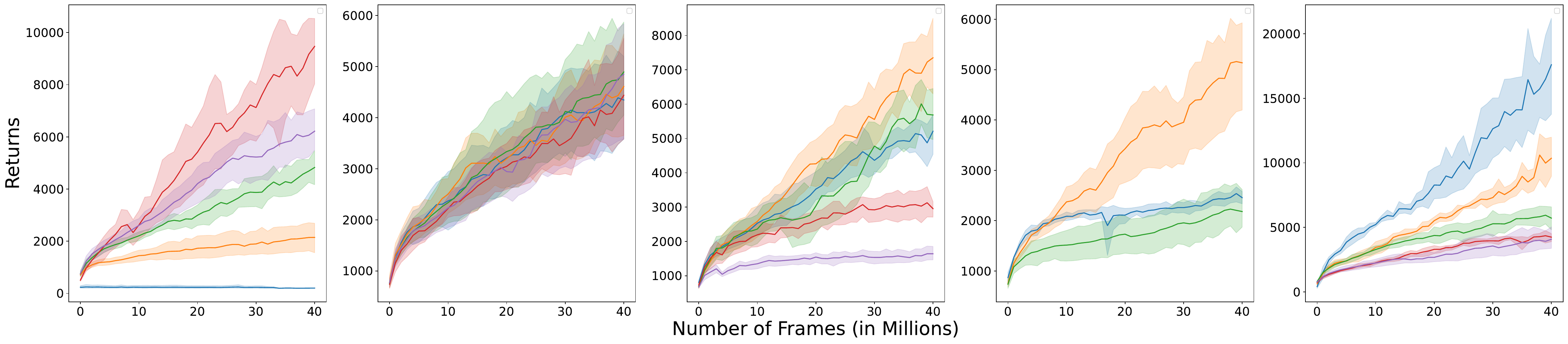}
  
    \caption{\textbf{Measured returns with various hyper-parameter variations on Asterix} for DrQ($\epsilon$) (top) and DER (bottom) at 40M environment frames. Displaying eight representative hyper-parameters, enabling per-game analyses for hyper-parameter selection.}%
    \label{fig:per_game}%
    \vspace{-1em}
\end{figure}

The website's main page presents aggregate IQM results for all hyper-parameters investigated in both data regimes (e.g. \autoref{fig:der_adam_eps}), while sub-pages present detailed performance comparisons when sliced by game (\autoref{fig:drq_eps_batch_sizes} presents a subset of this) and hyper-parameter (\autoref{fig:per_game} presents a subset of this).
The added level of granularity provided by the sub-pages can be crucial for understanding the specific strengths and weaknesses of an algorithm in various scenarios. All results averaged over 5 seeds, shaded areas represent 95\% confidence intervals.

\section{Related work}
\label{related_work}

While RL as a field has seen many innovations in the last
years, small changes to the algorithm or its implementation can have a big impact on its results \citep{engstrom2020implementation, joajo2021lifting}.  Deep reinforcement learning approaches are often notoriously sensitive to their hyperparamaters and demonstrate brittle convergence properties \citep{haarnoja2018soft}. This is particularly true for off-policy approaches that use a replay buffer to leverage past experiences \citep{duan2016benchmarking}.

\cite{Henderson2017DeepRL} investigate the effects of existing degrees of variability between various RL setups and their effects on algorithm performance. Although restricted to the domain of existing environments, \cite{Henderson2017DeepRL} propose more robust performance estimators for RL learning algorithms. \cite{islam2017reproducibility} and \cite{shengyi2022the37implementation} have shown the difficulty in reproducing policy gradient algorithms due to the variance.
\cite{andrychowicz2020matters} did a deep dive in algorithmic choices on policy-based algorithms. Their analyses covered differences in hyper-parameters, algorithms, and implementation details.

In an effort to consolidate innovations in deep RL, several papers have examined the effect of smaller design decisions like the loss function or policy regularization for on-policy algorithms \cite{andrychowicz2020matters}, DQN agents \citep{obando2020revisiting}, imitation learning \citep{hussenot2021hyperparameter} and offline RL \citep{paine2020hyperparameter, lu2021revisiting}. AutoRL methods, on the other hand, have focused on automating and abstracting some of these decisions \citep{parker2022automated, eimer2023hyperparameters} by using data-driven approaches to learn various algorithmic components or even entire RL algorithms \citep{co2021evolving,lu2022discovered}. All these works have demonstrated that hyperparameters in deep reinforcement learning warrant more attention from the research community than they currently receive. Underreported tuning practices can distort algorithm evaluations, and overlooked hyperparameters may lead to suboptimal performance.


\section{Discussion}
\label{sec:discussion}
One of the central challenges in reinforcement learning research is the non-stationarity during training in the inputs (due to self-collected data) and targets (due to bootstrapping). This is in direct contrast with supervised learning settings, where datasets and labels are typically fixed throughout training. This non-stationarity may be largely to blame for some of the ranking inconsistencies observed under different training regimes (e.g. \autoref{fig:der_adam_eps}), and why different hyper-parameter tunings are required for different settings (e.g. DER versus Rainbow).

Hyper-parameters are commonly tuned on a subset of environments (e.g. 3-5 games) and then evaluated on the full suite. Our findings suggest that this approach may not be the most rigorous, as hyper-parameter selection can vary dramatically from one game to the next (c.f. \cref{fig:drq_eps_batch_sizes,fig:per_game}).
While aggregate results (e.g. IQM) provide a succinct summary of performance, they unfortunately gloss over substantial differences in the individual environments. If our hope as researchers is to be able to use these algorithms beyond academic benchmarks, understanding these differences is {\em essential}, in particular in real-world applications such as healthcare and autonomous driving.

We have conducted a large number of experiments to investigate the impact of various hyper-parameter choices. While the THC score (\autoref{fig:this_score_all}) provides a high-level view of the transferability of hyper-parameter choices, our collective results suggest that a {\em single} set of hyper-parameter choices will never suffice to achieve strong performance across all environments. The ability to dynamically adjust hyper-parameter values during training is one way to address this; to properly do so would require quantifiable measures of environment characteristics that go beyond coarse specifications (such as sparse versus dense reward systems). The per-game results we present here may serve as an initial step in this direction. In Appendix~\ref{sec:gopher} we provide a fine-grained analysis of DER on Gopher as an example of the type of analyses enabled by our website. We hope our analyses, results, and website prove useful to RL researchers in developing robust and  transferable algorithms to handle increasingly complex problems.\\

\subsubsection*{Acknowledgements}

The authors would like to thank Jesse Farebrother, Gopeshh Subbaraj, Doina Precup, Hugo Larochelle, and the rest of the Google DeepMind Montreal team for valuable discussions during the preparation of this work.  Jesse Farebrother deserves a special mention for providing us valuable feed-back on an early draft of the paper. We thank the anonymous reviewers for their valuable help in improving our manuscript. We would also like to thank the Python community \cite{van1995python, 4160250} for developing tools that enabled this work, including NumPy \cite{harris2020array}, Matplotlib \cite{hunter2007matplotlib}, Jupyter \cite{2016ppap}, Pandas \cite{McKinney2013Python} and JAX \cite{bradbury2018jax}.

\subsubsection*{Broader Impact Statement}

Although the work presented here is mostly academic, it aids in the development of more capable and reliable autonomous agents. While our contributions do not directly contribute to any negative societal impacts, we urge the community to consider these when building on our research.

\bibliography{main}
\bibliographystyle{rlc}

\newpage
\appendix
\section{Computing Tuning Hyperparameter Consistency (THC)}
\label{sec:appendixTHC}
Computing the ranking between hyper-parameter values is non-trivial given the noise involved in Deep Reinforcement Learning Agents performances. We used 5 seeds to improve the robustness to noise of our results in this paper, but if we simply used the average performance the effects of noise would still be significant. As such our ranking is based on the Inter-Quantile Mean (IQM) \citep{agarwal2021deep} and its 95\% confidence interval.

First we sort the performance array in decreasing order based on the upper bound of the confidence interval for each hyper-parameter. Then we compute the rank of each hyper-parameter as the average between the lowest position (1-based) whose lower bound is less than or equal to the current hyper-parameter's performance upper bound and the highest position whose upper bound is greater than or equal to the current hyper-parameter's lower bound. Our choice of treating overlaps in performance by averaging the rankings comes from what is typically done when dealing with ties when computing Kendall's W and Kendall's $\tau$, which are other commonly used metrics for inter-ranking agreement.

As an example imagine we are analyzing a hyper-parameter with 5 possible values, 1e-2, 1e-1, 1, 1e1, 1e2. We run all the experiments and get the following confidence intervals on their IQM ranges $(200, 300), (250, 350), (400, 600), (110, 220), (30, 70)$. 
After sorting them we're left with: 
\begin{enumerate}
    \item 1: (400, 600)
    \item $10^{-1}$: (250, 350)
    \item $10^{-2}$: (200, 300)
    \item $10^1$: (110, 220)
    \item $10^2$: (30, 70)
\end{enumerate}
Then we can compute the ranks as: 
\begin{enumerate}
    \item $1: \frac{1 + 1}{2} = 1$
    \item $10^{-1}: \frac{2 + 3}{2} = 2.5$
    \item $10^{-2}: \frac{2 + 4}{2} = 3$
    \item $10^1: \frac{3 + 4}{2} = 3.5$
    \item $10^2: \frac{5 + 5}{2} = 5$
\end{enumerate}
An important feature of this method is that ranks needs not be integers. Now another relevant example is one where the 3 values, let's call them A, B, C, have completely overlapping intervals: \begin{enumerate}
    \item A: (200, 300)
    \item B: (250, 350)
    \item C: (180, 260)
\end{enumerate}
In this case all of them will have the ranking $\frac{1+3}{2} = 2$, which shows how given our results we're unable to fully determine which one is the best or worst performing value for this hyper-parameter. 

Here are two extra examples of computing the THC score.

\begin{enumerate}
    \item We analyze a case with 2 hyper-parameters, $A_1$ and $B_1$, both with 3 values, being evaluated across 5 games (columns are ranks in a game):

    \begin{equation*}
    \begin{aligned}
        r_{A_1} &= \begin{bmatrix}
                    1 & 1 & 2 & 1 & 3\\
                    2 & 3 & 2 & 3 & 2\\
                    3 & 2 & 2 & 2 & 1
                    \end{bmatrix} \\
        r_{B_1} &= \begin{bmatrix}
                    1 & 2 & 1 & 2 & 1\\
                    2 & 1 & 2 & 1 & 2\\
                    3 & 3 & 3 & 3 & 3
                    \end{bmatrix}
    \end{aligned}
    \xrightarrow{\text{peak-to-peak}}
    \begin{aligned}
        ptp_{A_1} &= \begin{bmatrix}
                    2\\
                    1\\
                    2
                    \end{bmatrix} \\
        ptp_{B_1} &= \begin{bmatrix}
                    1\\
                    1\\
                    0
                    \end{bmatrix}
    \end{aligned}
    \xrightarrow{\text{Normalize}}
    \begin{aligned}
        ptp_{A_1} &= \begin{bmatrix}
                    1.0\\
                    0.5\\
                    1.0
                    \end{bmatrix} \\
        ptp_{B_1} &= \begin{bmatrix}
                    0.5\\
                    0.5\\
                    0.0
                    \end{bmatrix}
    \end{aligned}
    \end{equation*}
    Finally we average the values to get the THC for each hyper-parameter:
        \begin{align}
            THC_{A_1} &= \frac{2.5}{3} \approx 0.83333  & THC_{B_1} &= \frac{1.0}{3} \approx 0.33333
        \end{align}

    This example also shows an important property of THC, while $a_1$ seems to be consistently the best value for A, whereas $b_1$ and $b_2$ vary their position more often, the value of THC is higher for A then for B, since the largest change in performance for values of A is larger than the change for values of B. This is because THC considers the worst-case variance when assigning how important is tuning a given hyper-parameter.
    
    \item Another example, now one hyper-parameter, $A_2$, has 4 possible values and the other, $B_2$, has 3, and we have 4 games.
    
\begin{equation*}
    \begin{aligned}
        r_{A_2} &= \begin{bmatrix}
                    1 & 1 & 1 & 3\\
                    2 & 2 & 2 & 2\\
                    3 & 3 & 3 & 1 \\
                    4 & 4 & 4 & 4
                    \end{bmatrix} \\
        r_{B_2} &= \begin{bmatrix}
                    1 & 1 & 1 & 1\\
                    2.5 & 2 & 3 & 2\\
                    2.5 & 3 & 2 & 3
                    \end{bmatrix}
    \end{aligned}
\xrightarrow{\text{peak-to-peak}}
\begin{aligned}
        ptp_{A_2} &= \begin{bmatrix}
                    2 \\
                    0 \\
                    2 \\
                    0
                    \end{bmatrix} \\
        ptp_{B_2} &= \begin{bmatrix}
                    0 \\
                    1 \\
                    1
                    \end{bmatrix}
    \end{aligned}
\xrightarrow{\text{Normalize}}
\begin{aligned}
        ptp_{A_2} &= \begin{bmatrix}
                    \frac{2}{3} \\
                    0.0 \\
                    \frac{2}{3} \\
                    0.0
                    \end{bmatrix} \\
        ptp_{B_2} &= \begin{bmatrix}
                    0.0\\
                    0.5\\
                    0.5
                    \end{bmatrix}
    \end{aligned}
\end{equation*}

    And then average across the hyper-parameter values:
        \begin{align}
            THC_{A_2} &= \frac{\frac{4}{3}}{4} = \frac{1}{3} & THC_{B_2} &= \frac{1}{3}
        \end{align}
    In this case we see that while $A_2$ has 2 hyper-parameter values with more variance in ranking then the 2 values of $B_2$ the fact that $A_2$ has more values overall than $B_2$ leads them to having the same THC value.
\end{enumerate}

Finally it's worth pointing out that since the performances in the second case were more stable than in the first one their THC value was overall lower. 

\section{Finer-grained experimental discussion}
\label{sec:finerGrainedExperiments}

\subsection{Optimal hyper-parameters do not Transfer Across Environments}
\begin{enumerate}
    \item For batch size in DrQ($\epsilon$)@40M we find that 4 is the optimal batch size for Asterix, Breakout, Gopher, and Seaquest, while being the worst value for effectively all the other games. See Figure \ref{fig:drq_eps_batch_sizes}
    \item Convolutional width for DER@40M, 0.25 is the clear optimum in Assault, CrazyClimber, Roadrunner, Seaquest, and UpNDown, while leading to the worst performance in Breakout, Krull, and QBert
    \item Dense layer width for DrQ($\epsilon$)@40M we see that 768 neurons per layer lead to best performance for Amidar, Assault, Hero, and Qbert, while most other games have 128 neurons as their optimal layer width. We see a similar mismatch for DER, though the games were 768 is optimal are different.
    \item A discount factor of 0.99 is optimal for DER@40M in Alien, Amidar, Asterix, BankHeist, Breakout, Frostbite, Kangaroo, Kung Fu Master, QBert, RoadRunner, Seaquest, and UpNDown, but leads to pessimal performance in PrivateEye and non-optmimum in Assault, Boxing, ChopperCommand, CrazyClimber and many others.
\end{enumerate}

\subsection{Optimal hyper-parameters do not Transfer Across Data Regimes}

\begin{enumerate}
    \item Adam's $\epsilon$, an often overlooked hyper-parameter, has optimal values < $1.5 \cdot 10^{-4}$ for Atari 100k, while having optimal value of $1.5 \cdot 10^{-2}$ in the 40M setting. This result also begs for further research, as higher values of $\epsilon$ move Adam closer to SGD with momentum behaviour.
    \item For Convolutional Width we find that the worst performing value for 100k, 0.25, is the optimal value when number of updates is 40M. Another important result given that it means one may want to effectively change the network architecture when the number of updates changes.
    \item For normalization of the dense layers we see that while in the 100k regime Layer Norm leads to worse performance than no normalization, it is the best performing normalization once we move to the 40M regime.
    \item For update horizon one can see that the best performing values are high, around 10, in the 100k regime, while lower values (as low as 1 for DER) are optimal in the 40M regime.
    \item For update period we see that in the 100k regime a value of 6 is low performing and 1 is optimal, but once we move to the 40M regime we see an inversion, where 6 is substantially superior to 1.
\end{enumerate}

\clearpage

\section{Hyper-parameters list}
\label{sec:list_hyperparameters}

Default hyper-parameter settings for DER and DrQ($\epsilon$) across the environments. \autoref{tbl:defaultvalues} shows the default values for each hyper-parameter across all the Atari games. In \autoref{tbl:allvalues} we list all the possible values we explored for both agents. The values selection is informed by the recommendations provided by \citet{joajo2021lifting}.

\begin{table}[!h]
 \centering
  \caption{Default hyper-parameters setting for DER and DrQ($\epsilon$) agents.}
  \label{tbl:defaultvalues}
 \begin{tabular}{@{} ccc @{}}
    \toprule
    & \multicolumn{2}{c}{Atari}\\
    \cmidrule(lr){2-3}
  Hyper-parameter &  DER & DrQ($\epsilon$) \\
    \midrule
     Adam's($\epsilon$) & 0.00015 & 0.00015\\
     Batch Size & 32 & 32\\
     Conv. Activation Function & ReLU & ReLU \\
     Convolutional Normalization & None & None \\
     Convolutional Width & 1& 1\\
     Dense Activation Function & ReLU & ReLU\\
     Dense Normalization & None & None \\
     Dense Width & 512 & 512 \\
     Discount Factor & 0.99 & 0.99 \\
     Exploration $\epsilon$ & & \\
     Learning Rate & 0.0001 & 0.0001 \\
     Minimum Replay History & & \\
     Number of Atoms & 51 & 0 \\
     Number of Convolutional Layers & & \\
     Number of Dense Layers & 2 & 2\\
     Replay Capacity & 1000000 & 1000000 \\
     Reward Clipping & True & True \\
     Update Horizon & 10 & 10 \\
     Update Period & 1& 1\\
    Weight Decay & 0 & 0\\
     \bottomrule
  \end{tabular}
\end{table}

\begin{table*}[!ht]
 \centering
  \caption{Hyper-parameters settings for DER and DrQ($\epsilon$) agents}
  \label{tbl:allvalues}
 \begin{tabular}{@{} cc @{}}
    \toprule
  Hyper-parameter &  Values \\
  \midrule
  Adam's($\epsilon$) & 1, 0.5, 0.3125, 0.03125, 0.003125, 0.0003125, 3.125e-05,\\
  & 3.125e-06 \\
  Batch Size & 4, 8, 16, 32, 64 \\
  Conv. Activation Function &  ReLU, ReLU6, Sigmoid, Softplus, Soft sign, SiLU, \\
  & Log Sigmoid, Hard Sigmoid, Hard SiLU, Hard tanh, ELU, \\
  & CELU, SELU, GELU, GLU \\
  Convolutional Normalization &  None, BatchNorm, LayerNorm \\
  Convolutional Width & 0.25, 0.5, 1, 2, 4 \\
  Dense Activation Function &  ReLU, ReLU6, Sigmoid, Softplus, Soft sign, SiLU, \\
  & Log Sigmoid, Hard Sigmoid, Hard SiLU, Hard tanh, ELU, \\
  & CELU, SELU, GELU, GLU \\
  Dense Normalization &  None, BatchNorm, LayerNorm \\
  Dense Width &  32, 64, 128, 256, 512, 1024 \\
  Discount Factor &  0.1, 0.5, 0.9, 0.99, 0.995, 0.999\\
  Exploration $\epsilon$ &  0, 0.001, 0.005, 0.01, 0.1\\
  Learning Rate & 10, 5, 2, 1, 0.1, 0.01, 0.001, 0.0001, 1e-05 \\
  Minimum Replay History & 125, 250, 375, 500, 625, 750, 875, 1000 \\
  Number of Atoms & 11, 21, 31, 41, 51, 61\\
  Number of Convolutional Layers &  1, 2, 3, 4\\
  Number of Dense Layers &  1, 2, 3, 4 \\
  Replay Capacity &  \\
  Reward Clipping &  True, False\\
  Target Update Period &  10, 25, 50, 100, 200, 400, 800, 1600\\
  Update Horizon & 1, 2, 3, 4, 5, 8, 10\\
  Update Period &  1, 2, 3, 4, 8, 10, 12\\
  Weight Decay &  0, 0.01, 0.03, 0.1, 0.5, 1\\

  \midrule
  \end{tabular}
\end{table*}

\clearpage

\section{Interesting Miscellaneous Findings}
\label{sec:imf}
There were a couple of interesting findings from our experiments which are out of scope for this paper, but which may warrant further exploration in the future.

\subsection{High Values of Weight Decay Can Be Optimal}
We found that for DER at 40 Million environment frames having a weight decay of 0.1 was the overall best choice, and that for many games like Gopher and Boxing the optimal value was 0.5, an uncommonly high value for the hyperparameter. 
\begin{figure}[!ht]
    \centering
  \includegraphics[width=0.6\linewidth]{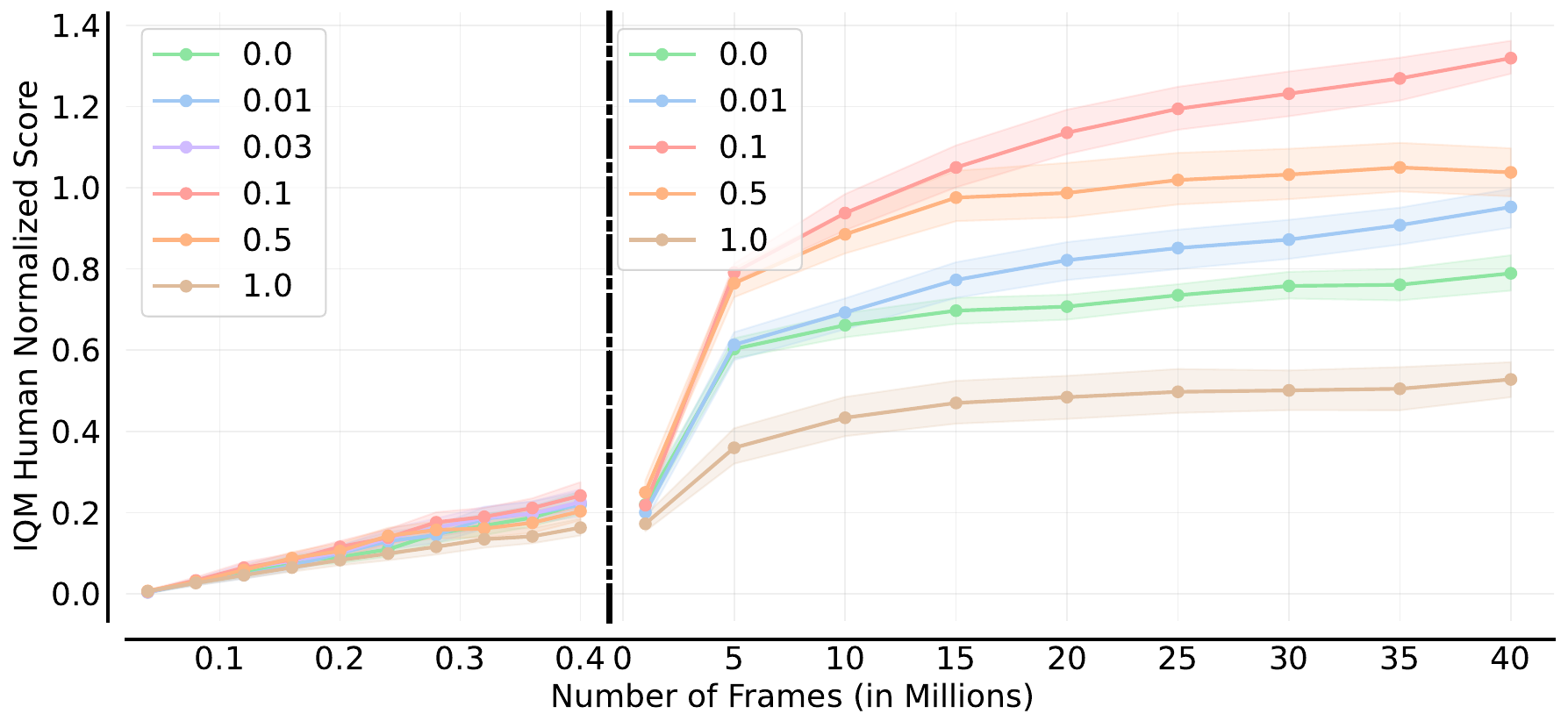}%
    \caption{
     \textbf{Measured IQM of human-normalized scores on the 26 100k benchmark games, with varying Weight Decay} for DER. We evaluate performance at 100k agent steps (or 400k environment frames), and at 40 million environment frames. At 40 million frames 0.1 is on average optimal, with 0.5 being at second place and the standard value of 0.0 being in fourth.
    }
    \label{fig:imf_wd}
\end{figure}

\subsection{Higher Values of Adam's $\epsilon$ can improve Performance}
In our experiments we found that both DrQ($\epsilon$) and DER can benefit from a 100 times higher value of Adam's $\epsilon$ than what is commonly used. This is somewhat perplexing, as using such a high value of epsilon leads Adam to behave closer to SGD than to its common behaviour in other settings.
\begin{figure}[!ht]
    \centering
  \includegraphics[width=0.8\linewidth]{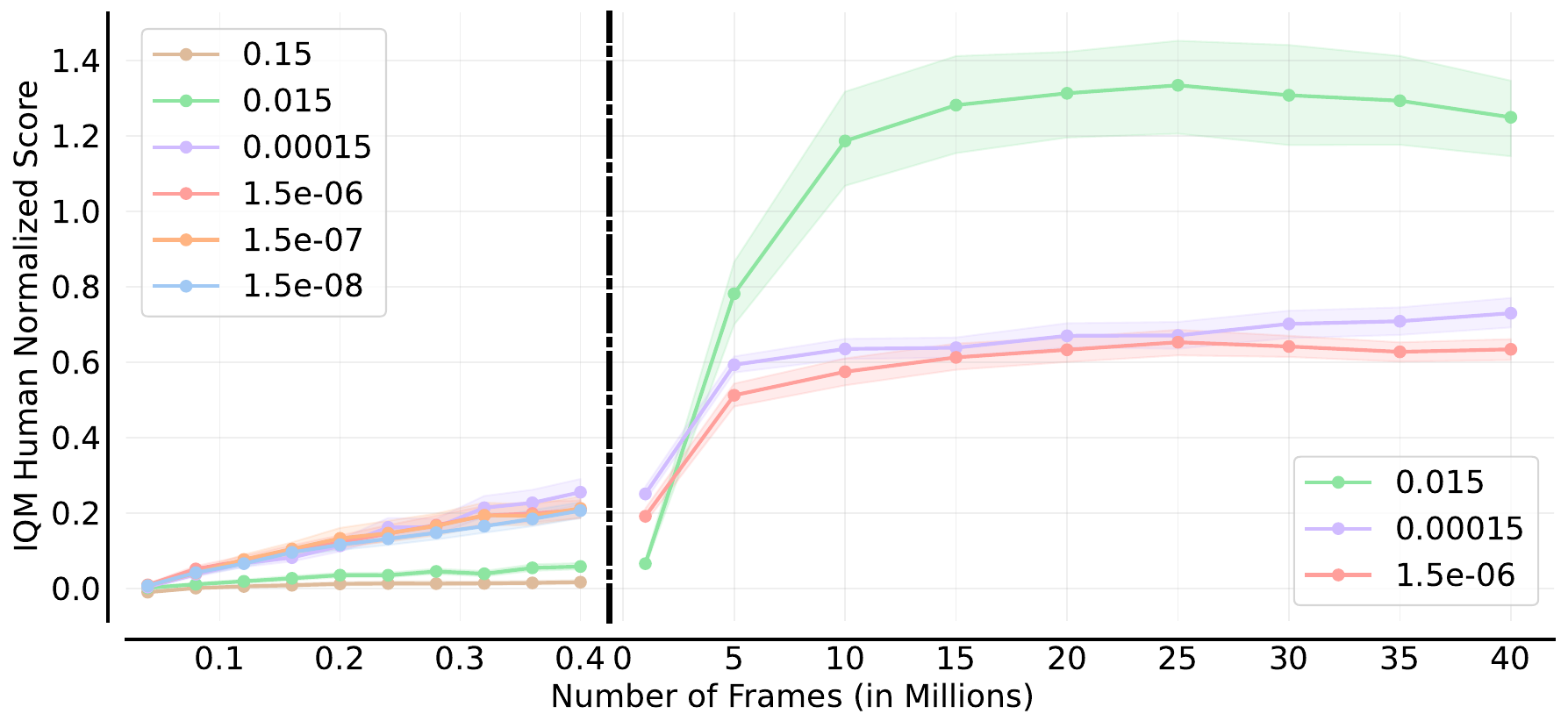}%
    \caption{
     \textbf{Measured IQM of human-normalized scores on the 26 100k benchmark games, with varying Adam's $\epsilon$} for DER. We evaluate performance at 100k agent steps (or 400k environment frames), and at 40 million environment frames.
    }
    \label{fig:imf_adam_eps}
\end{figure}

\subsection{Example Analysis: DER on Gopher}
\label{sec:gopher}
Finally we found that in a specific experimental setting, DER with 40 million frames on Gopher, whose optimal hyperparameters are very different from what is commonly observed in other applications of Deep Learning, and in some cases quite different even from the optimal values when using DER with 40 million environment frames in other Atari games. Not only that, but also we observed that often the difference in performance between the counter-intuitive optimal hyper-parameter and the standard is significant, leading to multiple-fold improvement in returns.
For example in Gopher specifically we find that:
\begin{itemize}
    \item For DER the standard value of update horizon is 10, but in the case of Gopher using an update horizon of 1 leads to roughly a 28 times improvement in performance.
    \item In Gopher a Weight Decay of 0.5 lead to a 5-fold increase of returns when compared to the standard value of 0.
    \item While the standard value of the Discount Factor is 0.99, for Gopher we see a 4.5 times improvement in performance when using a lower value of 0.9
    \item The optimal batch size we found was 4, which is relatively small compared to the standard of 32, and goes against the common Deep Learning practice of increasing batch sizes to increase performance. Changing batch size to 4 leads to a 4.5-fold increase in returns
    \item Finally, we recall the previous sub-section on Adam's $\epsilon$ and see that Gopher also benefits from an uncommonly high value of the hyperparameter, though here the performance gap is smaller, being closer to a 2x increase compared to the considerable differences discussed previously.
\end{itemize}

\begin{figure}[!ht]
    \centering
  \includegraphics[width=1.0\linewidth]{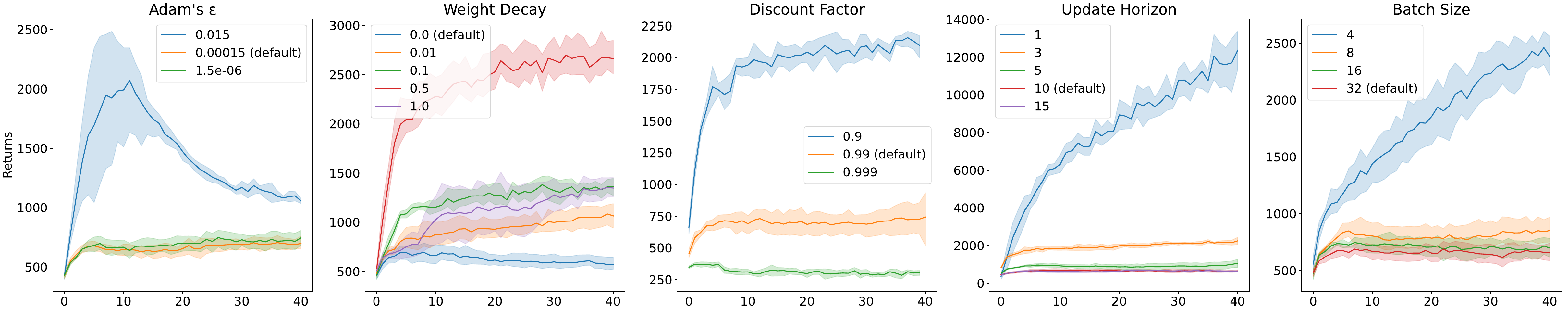}%
    \caption{
     \textbf{Learning Curves of DER on Gopher at 40M frames as we vary Adam's $\epsilon$, Weight Decay, Discount Factor, Update Horizon, and Batch Size}
    }
    \label{fig:weird_gopher}
\end{figure}

\end{document}